\documentclass[10pt,twocolumn,letterpaper]{article}

\usepackage[final]{iccv}      % To produce the REVIEW version
\usepackage{siunitx}
\usepackage[dvipsnames]{xcolor}

\newcommand{\figref}[1]{Fig.~\ref{#1}}
\newcommand{\secref}[1]{Sec.~\ref{#1}}

\newcommand{\tabref}[1]{Table~\ref{#1}}

\makeatletter
\DeclareRobustCommand\onedot{\futurelet\@let@token\@onedot}
\def\@onedot{\ifx\@let@token.\else.\null\fi\xspace}
\def\eg{e.g\onedot} 
\def\ie{i.e\onedot}

\def\wrt{wrt\onedot}

\def\etal{et~al\onedot}

\makeatother

\newcommand{\boldparagraph}[1]{\vspace{0.2cm}\noindent{\bf #1} }

\definecolor{darkgreen}{rgb}{0,0.7,0}
\definecolor{darkblue}{RGB}{31,119,180}
\definecolor{darkred}{RGB}{214,39,40}

\newcommand{\ourmodel}{GGS\xspace}
\newcommand{\ourstwod}{Ours-No3D\xspace}

\newcommand{\SUPP}{supplementary material\xspace}
\newcommand{\gs}{Gaussian splats\xspace}
\newcommand{\re}{RealEstate10K\xspace}
\newcommand{\snpp}{ScanNet++\xspace}
\newcommand{\sn}{ScanNet\xspace}

\newcommand{\singleViewScene}{
\begin{figure*}[t!]
\centering
  \includegraphics[width=\linewidth]{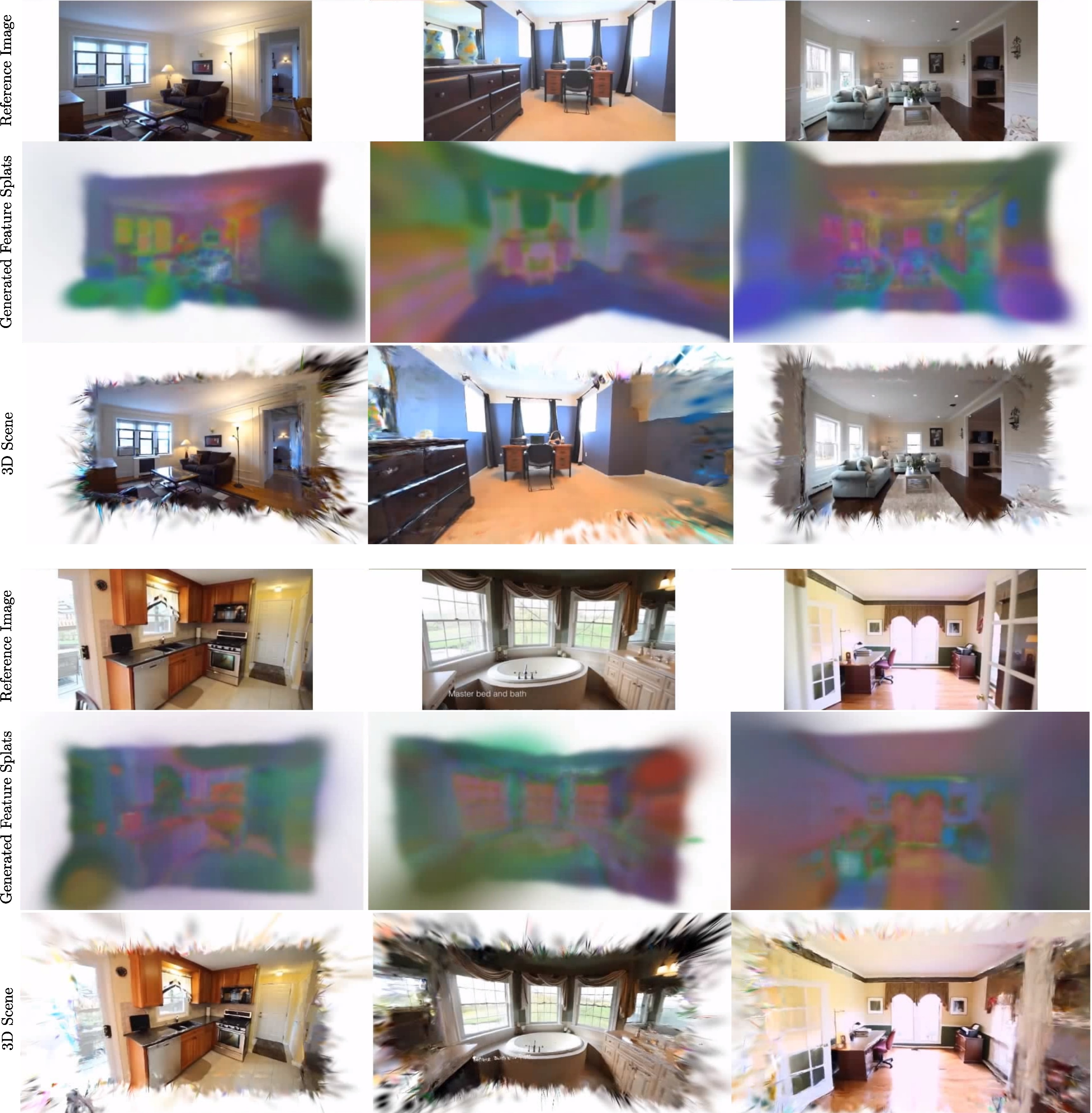}
  \caption{\textbf{3D Scene From a Single Image: }. We show generated \gs in image and feature space and the reference image.}
  \label{fig:singleViewScene}
\vspace{-1.5em}
\end{figure*}
}

\newcommand{\baselinecompfigSingle}{
    \begin{figure*}[t!]
      \centering
    \includegraphics[width=\linewidth]{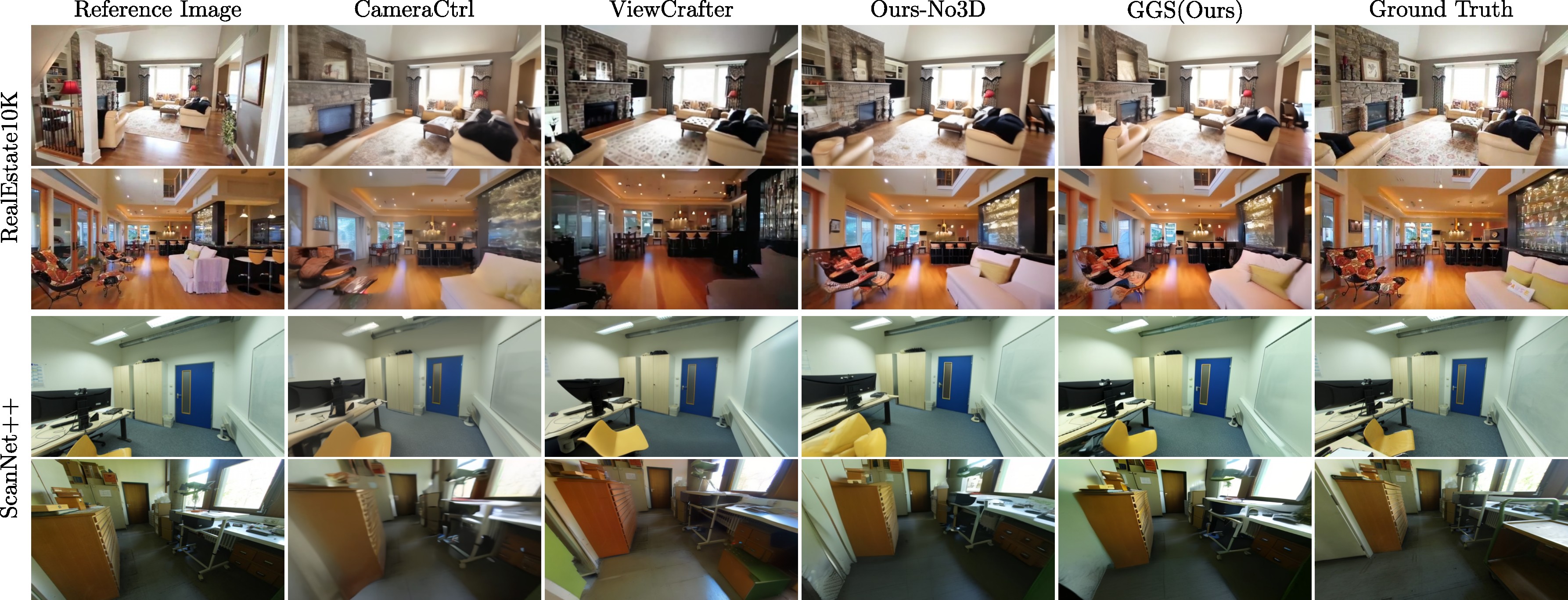}
      \caption{\textbf{Baseline Comparison Given One Reference Image: } We show results for the strongest baselines CameraCtrl~\cite{He2004CVPR} and ViewCrafter\cite{Yu2024ViewCrafter} together with our approach without (\ourstwod) and with 3D representation (\ourmodel). Best viewed zoomed in.}
      \label{fig:baselinecompfigSingle}
    \end{figure*}
}

\newcommand{\baselinecompfigSingleSupp}{
    \begin{figure*}[p]
      \centering
    \includegraphics[height=0.4\textheight]{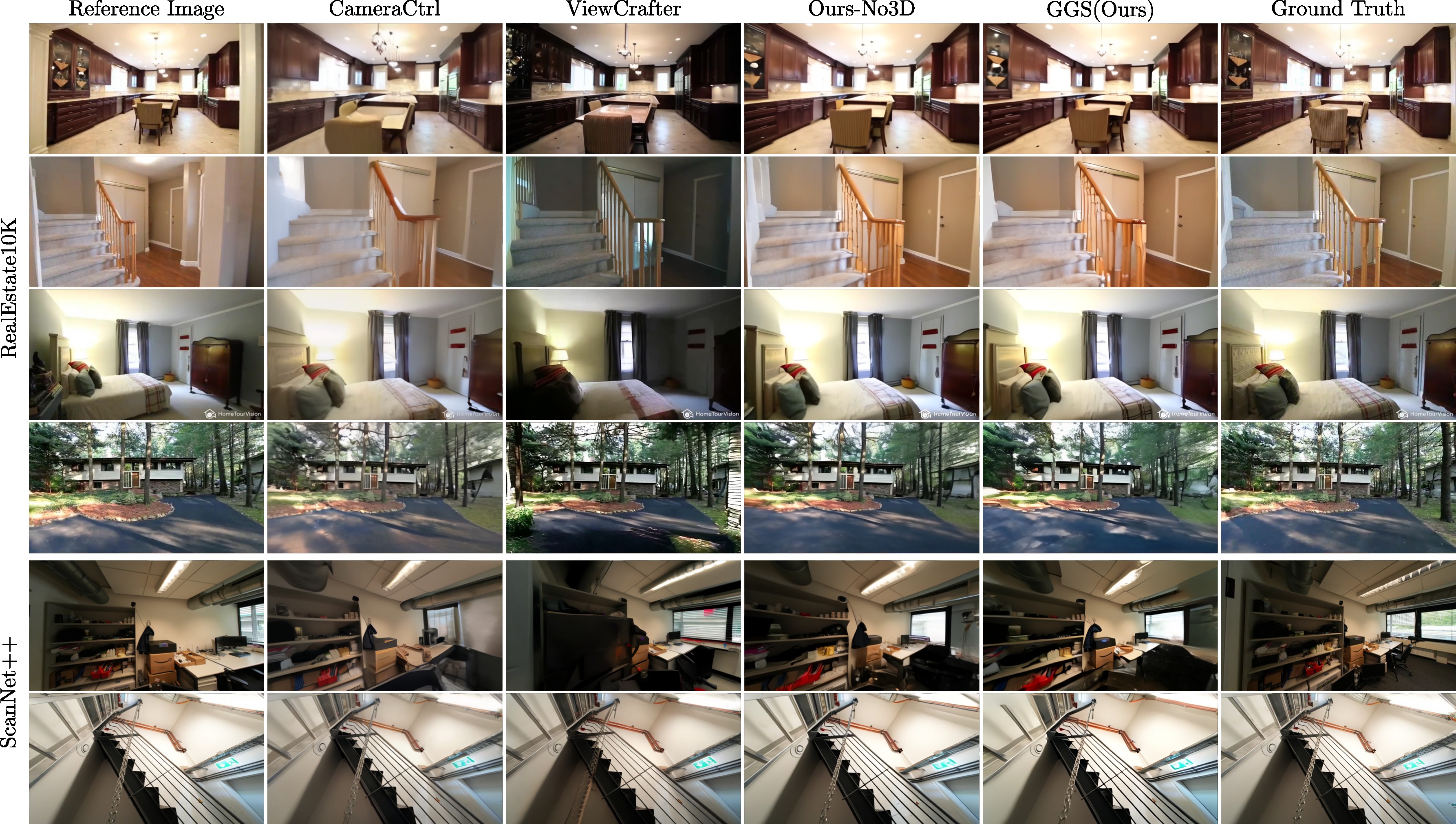}
      \caption{\textbf{Baseline Comparison Given One Reference Image: } We show results for the strongest baselines CameraCtrl~\cite{He2004CVPR} and ViewCrafter\cite{Yu2024ViewCrafter} together with our approach without (\ourstwod) and with 3D representation (\ourmodel). Best viewed zoomed in.}
      \label{fig:baselinecompfigSingleSupp}
\vspace{-1.5em}
    \end{figure*}
}

\newcommand{\baselinecompfig}{
    \begin{figure*}[t!]
      \centering
      \includegraphics[width=\linewidth]{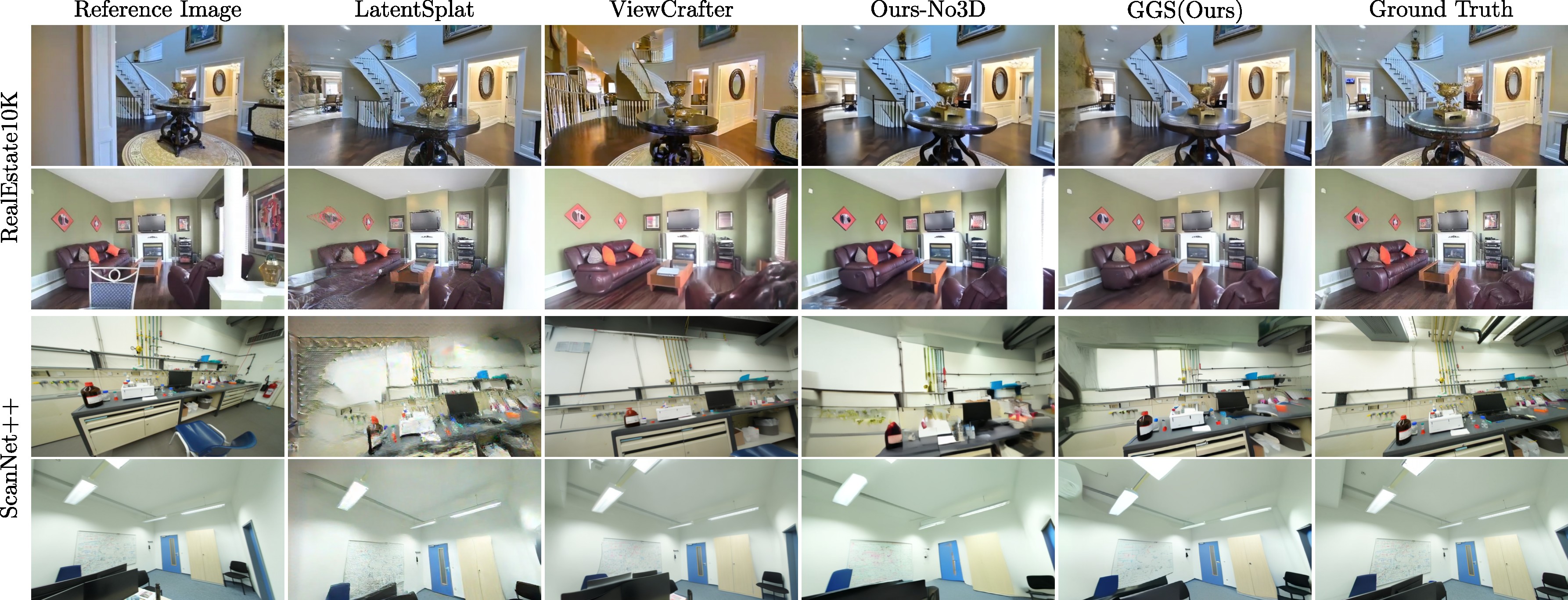}
      \caption{\textbf{Baseline Comparison For View Extrapolation Given Two Reference Images: } We show results for the strongest baselines LatentSplat~\cite{Wewer2024latentsplat} and ViewCrafter\cite{Yu2024ViewCrafter} together with our approach without (\ourstwod) and with 3D representation (\ourmodel). As both reference views are close together, we only include one image for reference. Best viewed zoomed in.}
      \label{fig:baselinecompfig}
      \vspace{-1.5em}
    \end{figure*}
}

\newcommand{\baselinecompfigSupp}{
    \begin{figure*}[p]
      \centering
      \includegraphics[height=0.4\textheight]{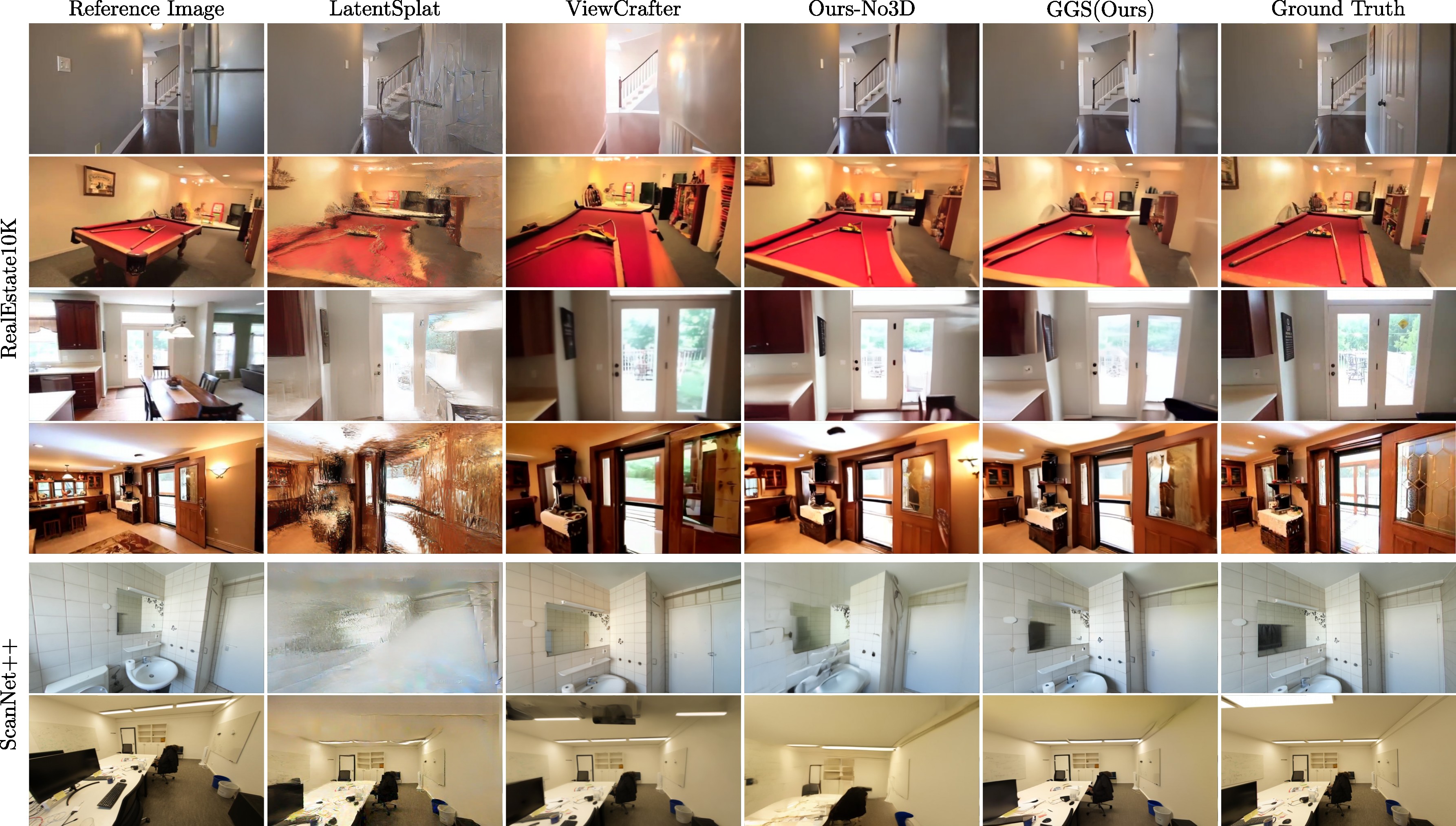}
      \caption{\textbf{Baseline Comparison For View Extrapolation Given Two Reference Images: } We show results for the strongest baselines LatentSplat~\cite{Wewer2024latentsplat} and ViewCrafter\cite{Yu2024ViewCrafter} together with our approach without (\ourstwod) and with 3D representation (\ourmodel). As both reference views are close together, we only include one image for reference. Best viewed zoomed in.}
      \label{fig:baselinecompfigSupp}
    \end{figure*}
}

\newcommand{\distillationComp}{
    \begin{figure*}[h!]
      \centering
    \includegraphics[width=\linewidth]{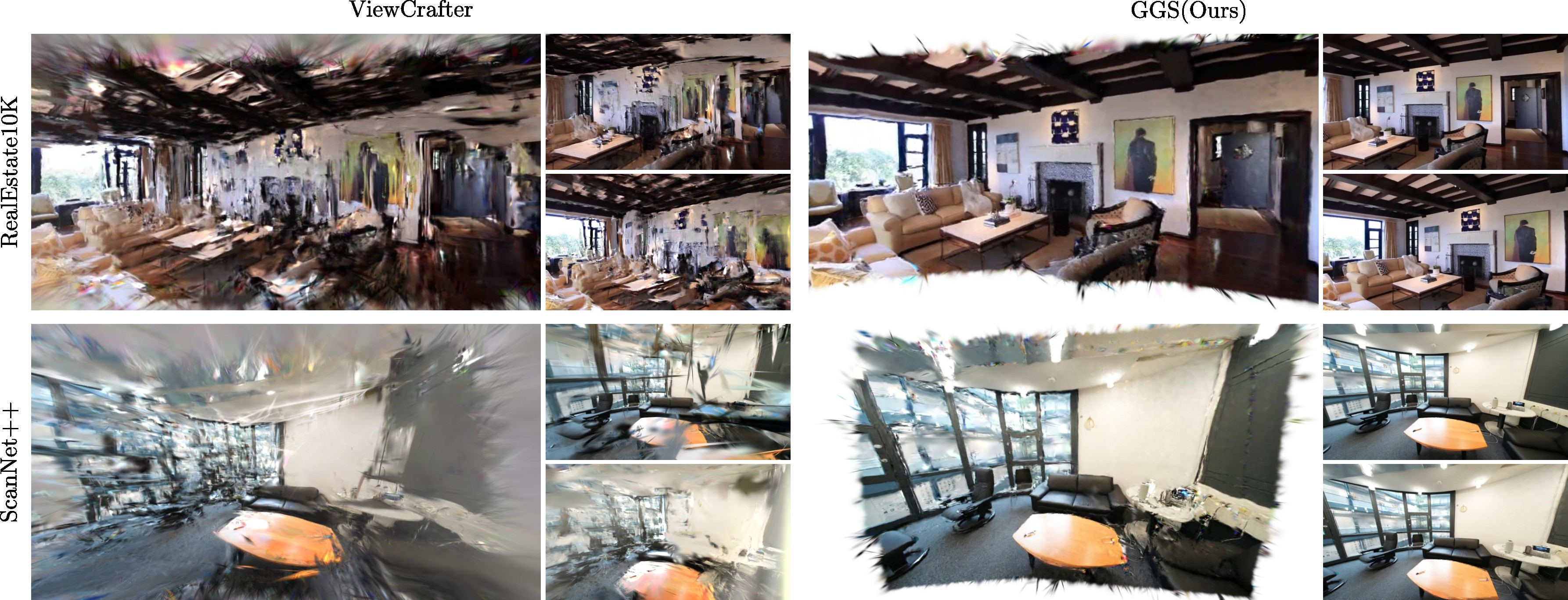}
    \caption{\textbf{3D Reconstruction Results From Generated Images: } We run an off-the-shelf 3DGS optimization on the generated multi-view images of ViewCrafter and \ourmodel(Ours). For ViewCrafter, we use 15,000 optimization steps. For our approach, we only refine the generated splats with the generated multi-view images, using 5,000 iterations. The resulting 3D representation is shown on the left and two rendered views from novel viewpoints are included on the right.
}
        \label{fig:distillation}
        \vspace{-1.5em}
    \end{figure*}
}

\newcommand{\distillationCompSupp}{
    \begin{figure*}[h!]
      \centering
    \includegraphics[width=\linewidth]{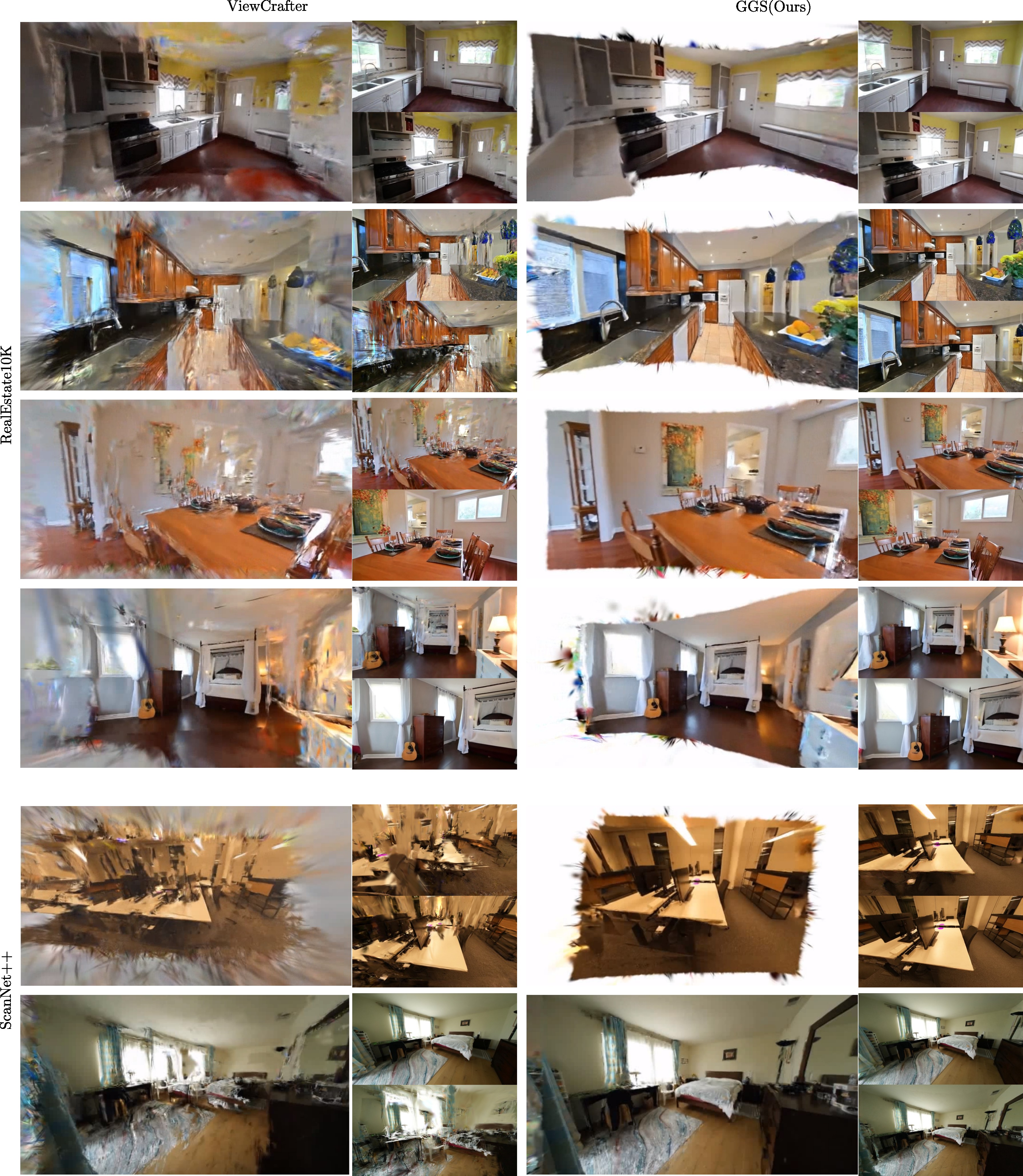}
    \caption{\textbf{3D Reconstruction Results From Generated Images: } We run an off-the-shelf 3DGS optimization on the generated multi-view images of ViewCrafter and \ourmodel(Ours). For ViewCrafter, we use 15,000 optimization steps. For our approach, we only refine the generated splats with the generated multi-view images, using 5,000 iterations. The resulting 3D representation is shown on the left and two rendered views from novel viewpoints are included on the right.
}
        \label{fig:distillationCompSupp}
        \vspace{-1.5em}

    \end{figure*}
}

\newcommand{\inpainting}{
    \begin{figure}[t!]
      \centering
        \includegraphics[width=\linewidth]{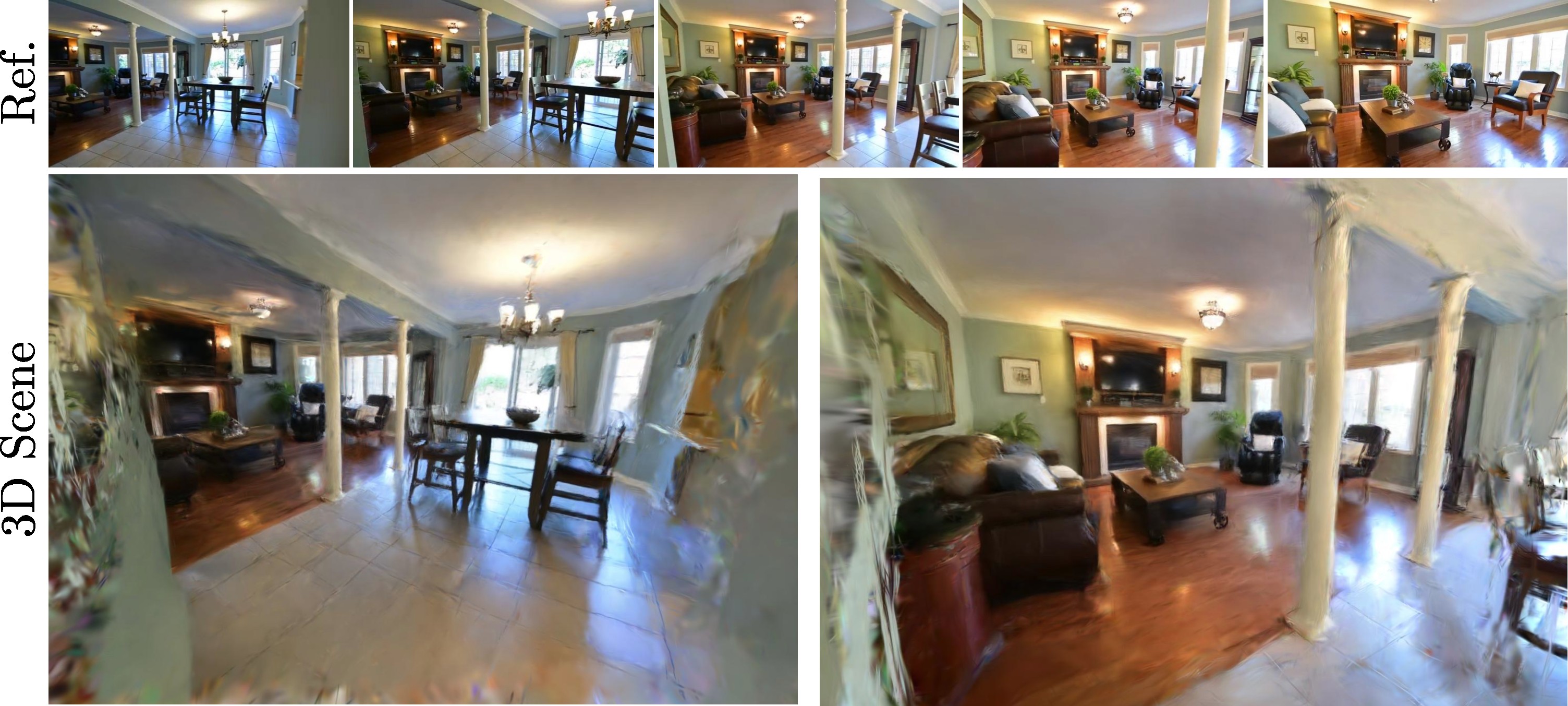}
      \caption{\textbf{Autoregressive Scene Synthesis with \ourmodel: } By generating consistent views between the reference images and from additional viewpoints, \ourmodel can augment the set of 5 reference images and generate larger 3D scenes autoregressively.}
      \label{fig:inpainting}
\vspace{-1.5em}
    \end{figure}
}

\newcommand{\inpaintingSupp}{
    \begin{figure*}[t!]
      \centering
        \includegraphics[width=\linewidth]{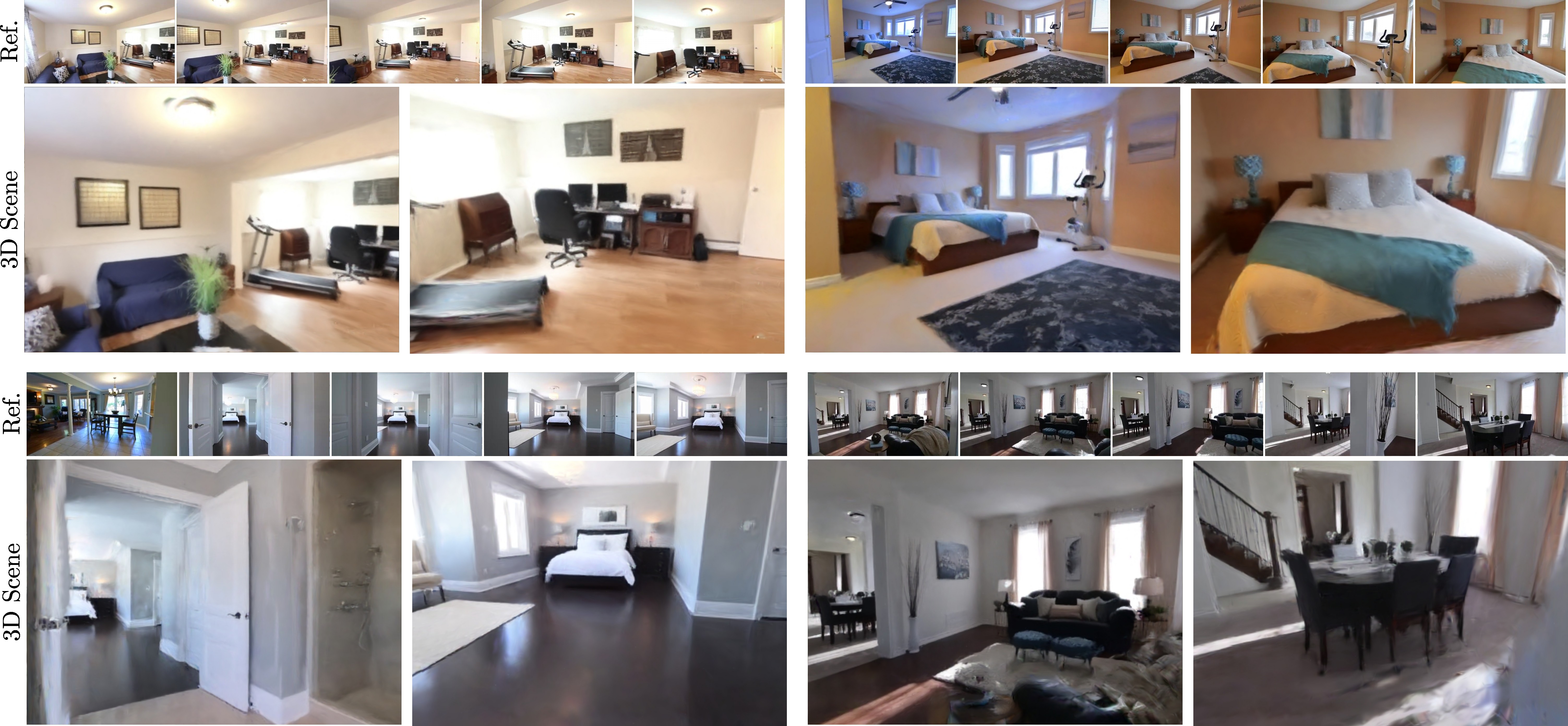}
      \caption{\textbf{Autoregressive Scene Synthesis with \ourmodel: } By generating consistent views between the reference images and from additional viewpoints, \ourmodel can augment the set of 5 reference images and generate larger 3D scenes autoregressively.}
      \label{fig:inpaintingSupp}
\vspace{-1.5em}
    \end{figure*}
}

\newcommand{\method}{
\begin{figure*}[t!]
\centering
  \includegraphics[width=0.99\textwidth]{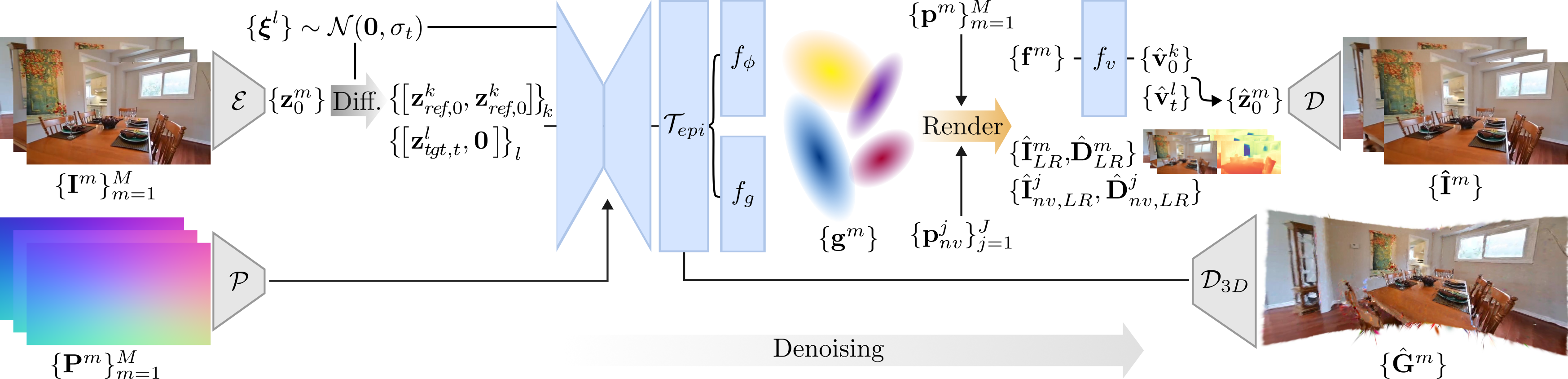}
  \caption{\textbf{Model Architecture: }Our approach, \ourmodel, directly synthesizes a 3D representation, which is parameterized by a set of Gaussian splats $\{\mathbf{g}^m\}$, from a set of posed input images. Specifically, during training we consider a set of posed images $\{\mathbf{I}^m\}$ with associated camera poses $\{\mathbf{p}^m\}$ and corresponding Pl\"ucker embeddings $\{\mathbf{P}^m\}$. The images are first encoded into a latent representation $\{\mathbf{z}_0^m\}$, which is then partitioned into $K$ reference images and $L$ target images. We introduce noise only to the latents of the target images $\{\mathbf{z}_{tgt,0}^l\}_{l=1}^L$, while leaving the reference images noise-free. To ensure compatibility with the pre-trained image-to-video diffusion model, we duplicate the reference latents across the channel dimension and concatenate zeros for the target latents. The resulting latents, along with the noise level $\sigma_t$ and Pl\"ucker embeddings, are fed into a U-Net architecture that produces intermediate per-latent feature maps. These feature maps are subsequently processed by an epipolar transformer $\mathcal{T}_{epi}$ to predict the parameters of the Gaussian feature splats $\{\mathbf{g}^m\}$. We render both feature maps $\{\mathbf{f}^m\}$ and low-resolution images $\{\mathbf{I}_{LR}^m\}$ for the input views, as well as low-resolution images for $J$ novel views $\{\mathbf{I}_{nv,LR}^j\}_{j=1}^J$ to regularize the 3D representation. Finally, the rendered feature maps are decoded into a weighted combination of sample noise $\mathbf{\xi}^m$ and input latent to predict the noise-free latents $\{\hat{\mathbf{z}}_0^m\}$.}
  \label{fig:method}
\vspace{-1.5em}
\end{figure*}
}
\newcolumntype{H}{>{\setbox0=\hbox\bgroup}c<{\egroup}@{}}       % hide columns by setting their type to H

\newcommand{\distillation}{
\begin{table}[t!]
    \centering
    \setlength{\tabcolsep}{12pt}
    \resizebox{\linewidth}{!}{
\begin{tabular}{lS[round-precision=1, table-format=2.1]S[round-precision=1, table-format=3.1]S[round-precision=1, table-format=2.1]S[round-precision=1, table-format=3.1]}\toprule
                                     & \multicolumn{2}{c}{RE10K}         & \multicolumn{2}{c}{ScanNet++}     \\
                                     &{FID$\downarrow$} & {FVD$\downarrow$} & {FID$\downarrow$} & {FVD$\downarrow$} \\\midrule
ViewCrafter                          &     70.80673839020572       &       784.6375862790813      &        119.76848796044978          &     553.3374682781725            \\
\ourstwod &    57.41162414230029            &       678.7181095364291        &      149.16364646589693         &      635.4932230930918      \\
\ourmodel(Ours)+depth &      57.36874978938545           &      \bfseries  440.5522303407957         &       127.64904572742321          &          523.8136385084828   \\
\ourmodel(Ours)+depth+FT &    \bfseries 47.567094802924146            &     468.3552594160796            &        \bfseries 119.28476138501925         &     \bfseries 513.7156834144389   \\\bottomrule    
\end{tabular}
    }
    \caption{\textbf{3D Scene Synthesis: } We report FID and FVD for rendered views between the training images at image resolution 576x320 pixels.}
    \label{tab:distillation}
\end{table}
}

\newcommand{\distillationSingle}{
\begin{table}[t!]
    \centering
    \setlength{\tabcolsep}{12pt}
    \resizebox{.8\linewidth}{!}{
\begin{tabular}{lS[round-precision=1, table-format=2.1]S[round-precision=1, table-format=3.1]S[round-precision=1, table-format=2.1]S[round-precision=1, table-format=3.1]}\toprule
                                     & \multicolumn{2}{c}{RE10K}         & \multicolumn{2}{c}{ScanNet++}     \\
                                     &{FID$\downarrow$} & {FVD$\downarrow$} & {FID$\downarrow$} & {FVD$\downarrow$} \\\midrule
ViewCrafter                          &     96.07774585030661       &       1071.836714283836      &        140.64551050735398          &     616.376318177019            \\
\ourstwod &    57.84421606339707            &      582.0289319494468       &      146.19437062431678         &       620.1046463254273      \\
\ourmodel(Ours)+depth &     56.3120420433058            &         498.2939485868205      &     127.4020392817266          &      587.8778372193923     \\
\ourmodel(Ours)+depth+FT &    \bfseries 47.32736692811045            &     \bfseries 403.86070294954925           &        \bfseries 110.99365382546128         &     \bfseries 548.0014775512644   \\\bottomrule    
\end{tabular}
    }
    \vspace{-1em}
    \caption{\textbf{Single Image to 3D: } FID and FVD scores for rendered views between the generated images at 576$\times$320 pixels.}
    \label{tab:distillationSingle}
    \vspace{-1.5em}
\end{table}
}

\newcommand{\ablations}{
\begin{table}[t!]
    \centering
    \setlength{\tabcolsep}{3pt}
    \resizebox{\linewidth}{!}{
\begin{tabular}{lS[round-precision=1, table-format=2.1]S[round-precision=3, table-format=1.3]S[round-precision=3, table-format=1.3]S[round-precision=1, table-format=2.1]S[round-precision=1, table-format=3.1]S[round-precision=3, table-format=1.3]}\toprule
& \multicolumn{3}{c}{Interpolation}                                                                                                                                                                                                                                                                                                                                                                            & \multicolumn{3}{c}{Extrapolation} \\
                                                   & {PSNR$\uparrow$} & {LPIPS$\downarrow$} & {TSED$\uparrow$} & {FID$\downarrow$} & {FVD$\downarrow$} & {TSED$\uparrow$} \\ \midrule
\ourmodel                              & 22.481233596801758        & 0.21437335014343262      & 0.9765625      &     51.31014923804357          &      316.99262685488566        &     0.971779336734694          \\
\quad No $\mathcal{L}_\text{nv,LR}$       &       19.60154151916504              &     0.2941841781139374      &      0.8683035714285713        &     65.8410905480323           &    320.8233809641824          &     0.8652742346938774          \\
\quad No $f_v$, $x_0$-prediction             &      14.879303932189941           &   0.6108666062355042       &     0.7528698979591837     &         239.401268811458        &   1131.5609288774506              &      0.821109693877551           \\\midrule
\ourmodel+depth  &       23.0457820892334         &    0.20426829159259796    &      0.9728954081632651        &      50.962948114835754        &        336.2607834534714         &         0.9631696428571427        \\
\quad $f_\phi$: Discrete $\phi$ &    21.9558048248291     &   0.2110145539045334        &    0.960140306122449   &      53.72877526742149      &      348.8906857147785            &       0.9564732142857143      \\\midrule
$\mathcal{D}_{3D}$: convolutional       &      21.160823822021484         &      0.2872922420501709          &      0.9953762755102041             &      66.9222702414192           &       317.9146574254094          &     0.989955357142857            \\
$\mathcal{D}_{3D}$: transformer         &       21.26308250427246              &  0.341801255941391       &      0.9961734693877551        &     92.2115361339122           &      343.531443270967          &       0.9896364795918365   \\\bottomrule
\end{tabular}
    }
    \caption{\textbf{Ablation Studies: }We investigate the effectiveness of our design choices on \re using two reference images.}
    \label{tab:ablations}
    \vspace{-1.5em}
\end{table}
}

\newcommand{\ablationsApp}{
\begin{table}[t!]
    \centering
    \setlength{\tabcolsep}{3pt}
    \resizebox{\linewidth}{!}{
\begin{tabular}{lS[round-precision=1, table-format=2.1]S[round-precision=3, table-format=1.3]S[round-precision=3, table-format=1.3]S[round-precision=1, table-format=2.1]S[round-precision=1, table-format=3.1]S[round-precision=3, table-format=1.3]}\toprule
& \multicolumn{3}{c}{Interpolation}                                                                                                                                                                                                                                                                                                                                                                            & \multicolumn{3}{c}{Extrapolation} \\
                                                   & {PSNR$\uparrow$} & {LPIPS$\downarrow$} & {TSED$\uparrow$} & {FID$\downarrow$} & {FVD$\downarrow$} & {TSED$\uparrow$} \\ \midrule
\ourmodel                              & 22.481233596801758        & 0.21437335014343262      & 0.9765625      &     51.31014923804357          &      316.99262685488566        &     0.971779336734694          \\
\quad No $\mathcal{L}_\text{nv,LR}$       &       19.60154151916504              &     0.2941841781139374      &      0.8683035714285713        &     65.8410905480323           &    320.8233809641824          &     0.8652742346938774          \\
\quad No $f_v$, $x_0$-prediction             &      14.879303932189941           &   0.6108666062355042       &     0.7528698979591837     &         239.401268811458        &   1131.5609288774506              &      0.821109693877551           \\\midrule
\ourmodel+depth  &       23.0457820892334         &    0.20426829159259796    &      0.9728954081632651        &      50.962948114835754        &        336.2607834534714         &         0.9631696428571427        \\
\quad $f_\phi$: Discrete $\phi$ &    21.9558048248291     &   0.2110145539045334        &    0.960140306122449   &      53.72877526742149      &      348.8906857147785            &       0.9564732142857143      \\\midrule
$\mathcal{D}_{3D}$: convolutional       &      21.160823822021484         &      0.2872922420501709          &      0.9953762755102041             &      66.9222702414192           &       317.9146574254094          &     0.989955357142857            \\
$\mathcal{D}_{3D}$: transformer         &       21.26308250427246              &  0.341801255941391       &      0.9961734693877551        &     92.2115361339122           &      343.531443270967          &       0.9896364795918365   \\
$\mathcal{D}_{3D}$: frozen backbone       &      19.5         &      0.312          &      0.942             &      70.1           &       320.7          &     0.934            \\\bottomrule
\end{tabular}
    }
    \caption{\textbf{Ablation Studies: }We investigate the effectiveness of our design choices on \re using two reference images.}
    \label{tab:ablationsApp}
    \vspace{-1.5em}
\end{table}
}

\newcommand{\baselinecompSingle}{
\begin{table}[t!]
    \centering
    \setlength{\tabcolsep}{3pt}
    \resizebox{\linewidth}{!}{
\begin{tabular}{lS[round-precision=1, table-format=2.1]S[round-precision=3, table-format=1.3]S[round-precision=1, table-format=2.1]S[round-precision=1, table-format=3.1]S[round-precision=3, table-format=1.3]S[round-precision=1, table-format=2.1]S[round-precision=3, table-format=1.3]S[round-precision=1, table-format=2.1]S[round-precision=1, table-format=3.1]S[round-precision=3, table-format=1.3]}\toprule
                                                         & \multicolumn{5}{c}{RE10K}                                                                               & \multicolumn{5}{c}{Scannet++}                                                                          \\
                                                         & {PSNR$\uparrow$}     & {LPIPS$\downarrow$}   & {FID$\downarrow$}    & {FVD$\downarrow$}    & {TSED$\uparrow$}    & {PSNR$\uparrow$}     & {LPIPS$\downarrow$}   & {FID$\downarrow$}   & {FVD$\downarrow$}    & {TSED$\uparrow$}    \\\midrule
PNVS                                                     &     16.658397674560547        &      0.3853469491004944        &   \bfseries    37.625396533563986        &       443.309997618146        &      0.9247448979591837     &         13.503582954406738           &        0.5164639949798584         &        102.87195184195095           &           497.9653218860034         &        0.36204081632653057            \\
MultiDiff                                                &       16.36606216430664          &          0.3714103698730469           &            43.72071346190304        &           443.309997618146          &       0.9432397959183673             &       14.292634010314941             &         0.4402333199977875            &          144.94881904764674         &      1130.17657671384              &      0.6828571428571429              \\
CameraCtrl                                               &       17.696073532104492            &           0.3323321044445038       &           46.20354655163506        &       366.66846476734395         &         0.8848852040816325           &   15.862235069274902       &       0.3990132808685303         &          82.00764413204641         &          516.646264231783        &         0.5004081632653061         \\
ViewCrafter                                              & 16.610855102539062             & 0.31236907839775085               & 44.201859054501924              & 313.2298686360987             & 0.9471131278360195              & 18.3159236907959              & \bfseries 0.2796415090560913              & \bfseries 74.52738511787535             & 437.52300689942854              & 0.8706122448979592            \\\midrule
\ourstwod                                              & 19.12198257446289 & \bfseries 0.2718009054660797 & 45.920945571389055 & 268.626299937473 & 0.983577806122449 & 17.09842872619629 & 0.35072240233421326 &  80.40379461768177 & \bfseries 436.3695487854057 & 0.5963265306122449 \\
\ourmodel (Ours)                               & \bfseries 19.195659637451172 & 0.2771002948284149 & 46.43357749706901 & \bfseries 267.4213980662023 & 0.9917091836734694 & 18.010942459106445 & 0.3170269727706909 & 81.75226057954886 & 455.4574647149669 & 0.8955102040816326 \\
\ourmodel(Ours)+depth                            & \bfseries 19.196062088012695 & 0.27268821001052856 & 46.989176233692774 & 273.54540798828424 & \bfseries 0.9992028061224489 & \bfseries 18.384435653686523 &  0.309450626373291 & 81.91292072487943 & 441.8517260243023 & \bfseries 0.8987755102040816 \\\bottomrule
\end{tabular}
    }
    \caption{\textbf{Baseline Comparison Given One Reference Image: } We benchmark the approaches on \re and evaluate generalization ability on \snpp. The reported metrics are calculated on all generated images.}
    \label{tab:baselinecompSingle}
    \vspace{-1.5em}
\end{table}
}

\newcommand{\baselinecomp}{
\begin{table}[t!]
    \centering
    \setlength{\tabcolsep}{3pt}
    \resizebox{\linewidth }{!}{
\begin{tabular}{l
S[round-precision=1, table-format=2.1]S[round-precision=3, table-format=1.3]HHS[round-precision=3, table-format=1.3]HHS[round-precision=1, table-format=2.1]S[round-precision=1, table-format=3.1]S[round-precision=3, table-format=1.3]
S[round-precision=1, table-format=2.1]S[round-precision=3, table-format=1.3]HHS[round-precision=3, table-format=1.3]HHS[round-precision=1, table-format=2.1]S[round-precision=1, table-format=3.1]S[round-precision=3, table-format=1.3]
}\toprule
                                                         & \multicolumn{10}{c}{RE10K}                                                                                                                                                                                                                                                                                                                                                                            & \multicolumn{10}{c}{Scannet++}                                                                                                                                                                                                                                                                                                                                             \\
                                                         & \multicolumn{5}{c}{Interpolation}                                                                                                                                                    & \multicolumn{5}{c}{Extrapolation}                                                                                                                                                                              & \multicolumn{5}{c}{Interpolation}                                                                                                                           & \multicolumn{5}{c}{Extrapolation}                                                                                                                                                                            \\
                                                         & {PSNR$\uparrow$}     & {LPIPS$\downarrow$}   & {FID$\downarrow$}    & {FVD$\downarrow$}    & {TSED$\uparrow$}    & {PSNR$\uparrow$}     & {LPIPS$\downarrow$}   & {FID$\downarrow$}   & {FVD$\downarrow$}    & {TSED$\uparrow$}    
                                                         & {PSNR$\uparrow$}     & {LPIPS$\downarrow$}   & {FID$\downarrow$}    & {FVD$\downarrow$}    & {TSED$\uparrow$}    & {PSNR$\uparrow$}     & {LPIPS$\downarrow$}   & {FID$\downarrow$}   & {FVD$\downarrow$}    & {TSED$\uparrow$}    \\\midrule                                                         
PixelSplat                                               & \bfseries 23.86610984802246                                     & \bfseries 0.170102059841156                                     &              &                    & \bfseries 1.0                                         &                   \text{--}                       &                           \text{--}               &     \text{--}                                      &\text{--}                    &\text{--}                                         &        21.158632278442383            &            0.2276461273431778         &\text{--}                                        &\text{--}                    &                      \bfseries 0.9485714285714285                  &                                             &                                             &\text{--}                                       &\text{--}                    &\text{--}                                         \\
LatentSplat                                              & 22.138757705688477                                     & 0.19482962787151337                                       &                    &                        &               \bfseries      1.0                         & 18.623271942138672                                       & 0.3288060426712036                                      &              59.766457056065306                                &             428.6147225449763          &                         0.9567920918367347                     &           \bfseries 21.690364837646484         &             0.27355098724365234        &                                            &                        &                        0.9191836734693878                    &                18.78767967224121                             &                0.37981942296028137                             &                    131.37257446017466                       &       638.3845619037918               &                        0.8395918367346937                    \\
ViewCrafter                                &                  19.850669860839844                         &                    0.2303393930196762                       &          45.757363274778754          &                        &                   0.8977997448979592                         &            17.165456771850586                                 &       0.31699272990226746                                      &                     \bfseries 48.477750062621176                      &         319.997095657338               &                       0.9001913265306124                      & 20.377737045288086 & \bfseries  0.2175265997648239 &                             &      & 0.8375510204081632                         & 19.1389026641845                        & 0.26512762904167175                         & \bfseries 80.99241244475323                        & \bfseries 473.1890916882072      & 0.8444897959183674                          \\ \midrule
\ourstwod & 21.5701961517334                           & 0.21414364874362946                         & 48.50716956023996  & \text{--} & 0.9162946428571429                          & 18.245790481567383                          & 0.3231874704360962                          & 52.662752780983915                          & 306.2171087678739 & 0.8920599489795917                        & 18.01677703857422  & 0.31677016615867615 & \bfseries 84.56143732589453 & \text{--} & 0.5322448979591837                         & 17.14962387084961                           & 0.359015554189682                           & 87.53147726349954                         & 510.98538010479126 & 0.5951020408163264     \\                    
\ourmodel(Ours)          & 22.866539001464844                         & 0.20399601757526398                          &  &  & 0.9701849489795916                          &                          &                           & 52.61872528281287                         & 324.8512387987238 & 0.9743303571428571                          & 19.286975860595703 &0.30342423915863037  &                         &  &  0.746938775510204                         &                           &                          & 98.63386911060081                   & 497.0266417283125 & 0.7938775510204081                           \\
\ourmodel(Ours)+depth   & 23.669204711914062                         & 0.18940524756908417                         & 48.3713190551545   & \text{--} & 0.9829400510204083 & 20.555097579956055                          & 0.26029372215270996                         & 49.426030476592494 & \bfseries 293.8484749044927 & \bfseries 0.9897959183673469 & 20.15415382385254  & 0.2660207748413086  & 87.75818472502704                          & \text{--} &  0.843265306122449 & \bfseries 20.450511932373047 & \bfseries 0.2558344602584839 & 85.6452676601153 & 476.6490212747246 & \bfseries 0.8542857142857143 \\\bottomrule
\end{tabular}
    }
    \caption{\textbf{Baseline Comparison Given Two Reference Images: } We benchmark the approaches on \re and evaluate generalization on \snpp. We report PSNR, LPIPS and TSED for view interpolation, and FID, FVD and TSED for view extrapolation. The metrics are calculated on all generated images.}
    \label{tab:baselinecomp}
    \vspace{-1.5em}
\end{table}
}

\definecolor{iccvblue}{rgb}{0.21,0.49,0.74}
\usepackage[pagebackref,breaklinks,colorlinks,allcolors=iccvblue]{hyperref}

\def\paperID{11305} % *** Enter the Paper ID here
\def\confName{ICCV}
\def\confYear{2025}

\title{Generative Gaussian Splatting: \\Generating 3D Scenes with Video Diffusion Priors}

\author{Katja Schwarz\quad Norman M\"uller\quad Peter Kontschieder\vspace{1em}\\
Meta Reality Labs Zurich, Switzerland
}

\begin{document}
\twocolumn[{
        \renewcommand\twocolumn[1][]{#1}%
    \maketitle
        %\vspace{-1cm}
    \begin{center}
    \includegraphics[width=\linewidth, trim={0 0.0cm 0.0cm 0cm},clip]{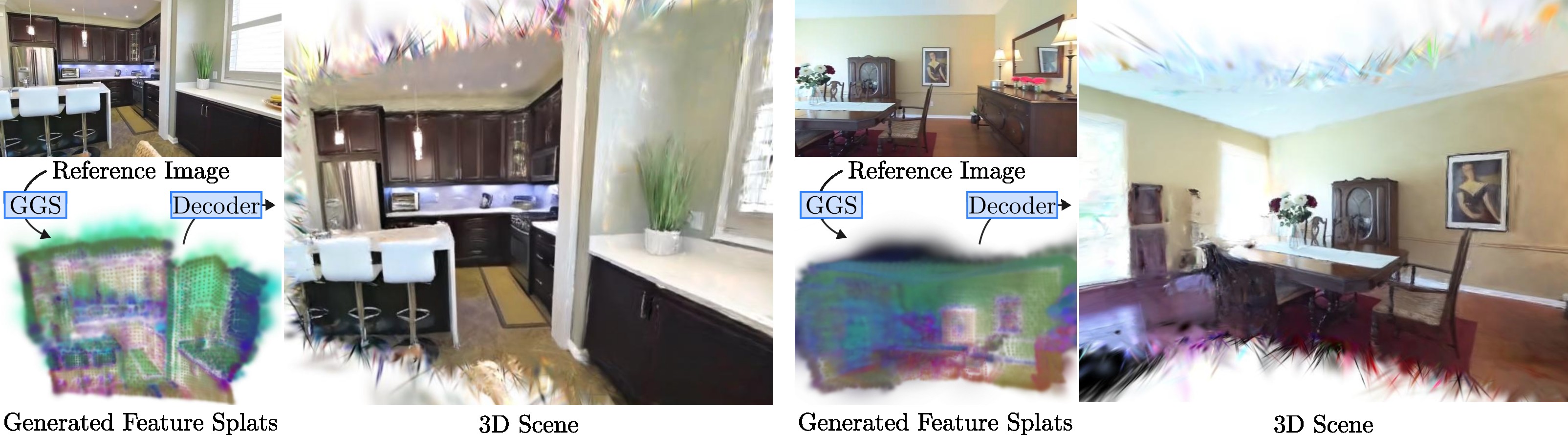}
         \captionof{figure}{\textbf{Overview: }Given one or more input images, \ourmodel leverages a video diffusion prior to directly generate a 3D radiance field parameterized via 3D Gaussian primitives. \ourmodel first generates a feature field with a pose-conditional diffusion model and subsequently decodes the feature splats, yielding an explicit 3D representation of the generated scene. Project page: \url{https://katjaschwarz.github.io/ggs/}
            }
        \label{fig:teaser}
    \end{center}    
}]

\begin{abstract}
\noindent Synthesizing consistent and photorealistic 3D scenes is an open problem in computer vision. Video diffusion models generate impressive videos but cannot directly synthesize 3D representations, \ie, lack 3D consistency in the generated sequences. In addition, directly training generative 3D models is challenging due to a lack of 3D training data at scale. In this work, we present Generative Gaussian Splatting (\ourmodel) -- a novel approach that integrates a 3D representation with a pre-trained latent video diffusion model. Specifically, our model synthesizes a feature field parameterized via 3D Gaussian primitives. The feature field is then either rendered to feature maps and decoded into multi-view images, or directly upsampled into a 3D radiance field.
We evaluate our approach on two common benchmark datasets for scene synthesis, \re and \snpp, and find that our proposed \ourmodel model significantly improves both the 3D consistency of the generated multi-view images, and the quality of the generated 3D scenes over all relevant baselines. Compared to a similar model without 3D representation, \ourmodel improves FID on the generated 3D scenes by $\sim$20\% on both \re and \snpp.
\end{abstract}
    
\section{Introduction}
Diffusion models work remarkably well for generating photorealistic images and videos from noise when trained on vast amounts of data. For 3D scenes, however, 3D training data is scarce and generative models fall short in terms of quality and generalization ability compared to their 2D counterparts. As a consequence, some recent works leverage pre-trained video diffusion models as backbones to first generate multi-view images, and subsequently stitch the generated views together using 3D reconstruction algorithms~\cite{wu2023reconfusion,Yu2024ViewCrafter,Gao2024Cat3d}. However, the generated multi-view images often lack 3D consistency, requiring carefully tailored 3D reconstruction algorithms~\cite{wu2023reconfusion,Gao2024Cat3d} or time consuming iterative procedures~\cite{Yu2024ViewCrafter}.\\
Another line of works improve 3D consistency of the generated images by using a 3D representation within the diffusion model~\cite{Anciukevicius2024ImageBasedDiffusion,tewari2023diffusion,chan2023genvs}. However, these works cannot leverage pre-trained video diffusion models, because of their custom network architectures for incorporating the 3D representation. In this work, we investigate how 3D representations can be directly integrated with powerful video diffusion priors to improve the consistency of the generated images and thereby the generated 3D scenes. One challenge is that state-of-the-art diffusion models operate on a compressed latent space, which is spatially approximately aligned with the input images but itself is not 3D-consistent. Another challenge is that predicting noise instead of the denoised input in practice works better and is the de-facto standard in video diffusion models. However, when including a 3D representation into diffusion models, this representation should mirror the denoised input, \ie the 3D scene, and cannot directly model the statistically independent noise added to the input images.\\
We find that we can solve both aforementioned issues by learning a 3D representation in feature space instead of latent space (see Figure~\ref{fig:teaser}). In particular, our Generative Gaussian Splatting (\ourmodel) approach renders feature maps from the 3D representation that are subsequently decoded to latent space. This allows for enough flexibility to train with $v$-prediction while enabling the training procedure to generate a meaningful 3D representation. Since this refinement can again introduce inconsistencies in the generated images, we additionally propose a decoder that directly predicts a decoded 3D scene from the generated feature maps.\\
Our experiments demonstrate that including a 3D representation indeed substantially improves the consistency across generated images. Similarly, our generated 3D scenes are inherently 3D-consistent, and their quality can be further %show improved quality, which can be further 
refined by \eg running a few iterations of a standard 3D reconstruction algorithm together with our generated 2D images. Compared to a similar model without 3D representation, \ourmodel improves FID on the generated 3D scenes by $\sim$20\%, see \tabref{tab:distillationSingle}. Another interesting property of our approach is that using an explicit 3D representation like \gs %, our approach also na\"ively 
supports training with additional depth supervision where available, resulting in additional improvements, particularly in terms of consistency.
We summarize our main contributions as follows:
\begin{itemize}
    \item We propose an approach that directly integrates an explicit 3D representation with a pre-trained latent video diffusion backbone, thereby improving 3D consistency of the generated image sequences and allowing for training with additional depth supervision where available.
    \item We design a custom decoder that directly predicts the decoded 3D representation of the scene from the generated feature maps.
    \item We train a conditional variant of our model that auto-regressively generates full scenes from an arbitrary number of input views.
\end{itemize}

\section{Related Works}
\label{sec:relworks}
\boldparagraph{Regression-Based Models for Novel View Synthesis.}
Recently, multiple works consider NVS from sparse input views, leveraging priors learned across many training scenes~\cite{yu2021pixelnerf,sitzmann2019scene,niemeyer2022,niemeyer2020dvr,wang2021ibrnet,Henzler_2021_CVPR,du2023wide,chen2021mvsnerf,trevithick2020GRF,sajjadi2022scene,kulhanek2022viewformer,roessle2022depthpriorsnerf,Hong2024LRM,Zhang2024GSLRM,Charatan2024PixelSplat,Wewer2024latentsplat,chen2024mvsplat,Fan2024instantsplat,Zhu2024FSGS,Szymanowicz2024SplatterImage,jin2024lvsm}. PixelSplat~\cite{Charatan2024PixelSplat}, Splatter Image~\cite{Szymanowicz2024SplatterImage}, and GS-LRM~\cite{Zhang2024GSLRM} directly predict per-pixel \gs. All of these methods optimize a regression objective, and hence cannot perform longer-range extrapolations. Instead, we consider a generative model to enable view extrapolation. Similarly, LatentSplat~\cite{Wewer2024latentsplat} addresses this limitation by extending PixelSplat with a GAN-based decoder, enabling moderate-distance extrapolations. In contrast to LatentSplat, our approach leverages a strong generative backbone and thereby supports larger view extrapolation. 

\boldparagraph{3D Generative Models.}
A few works directly apply generative models to 3D representations~\cite{Zeng2022lion,Luo2021dpmpointcloud,Zhou2021pointvoxeldiffusion} but the lack of 3D training data makes it difficult to train such models at large scale. Thus, many works combine an intermediate 3D representation with differentiable rendering and train with posed images instead of 3D data~\cite{chan2023genvs,Schwarz2024wildfusion,schwarzvoxgraf,schwarz2020graf,bautista2022gaudi,ren2024scube,tewari2023diffusion,Anciukevicius2024ImageBasedDiffusion,kim2023nfldm,niemeyer2021giraffe}. However, compared to images and video data, posed multi-view images are still difficult to obtain at large scale. Consequently, 3D generative models do not yet match in quality and generalization ability \wrt to their 2D counterparts. The aforementioned methods require custom architectures for incorporating the 3D representation and can hence not leverage pre-trained backbones. In contrast, we combine a 3D representation with a pre-trained video diffusion model that acts as a powerful prior. Concurrently, Prometheus~\cite{yang2024prometheus} trains a text-to-3D diffusion model by learning to denoise depth and multi-view images jointly.
\paragraph{Pose-Conditional Image and Video Diffusion.}
Diffusion Models (DMs)~\cite{ho2020denoising,sohldickstein2015deep,song2020score} achieve state-of-the-art results in text- and image-guided synthesis~\cite{nichol2021improved,rombach2021highresolution,dhariwal2021diffusion,ho2022cascaded,podell2023sdxl,ho2022imagenvideo,singer2022make,luo2023videofusion,esser2023structure}. 
Recent works propose pose-conditional variants~\cite{Liu2024one2345pp,Watson20244DiM,wu2023reconfusion,Gao2024Cat3d,Yu2024ViewCrafter,Shriram2024realmdreamer,Liang2024luciddreamer,poole2022dreamfusion,Shi2024mvdream,Voleti2024SV3D,watson20233dim,liu2023zero1to3,Xu2024Camco,He2024CameraCtrl,Wang2024MotionCtrl,sargent2023zeronvs,zhou2023sparsefusion,Yu2023PhotoconsistentNVS,liang2024wonderland,sun2024dimensionx}. PNVS~\cite{Yu2023PhotoconsistentNVS} trains a 2D diffusion model to autoregressively generate frames along a given camera trajectory. MultiDiff~\cite{Muller2024MultiDiff} conditions the DM on depth-based warped images. More recently, CAT3D~\cite{Gao2024Cat3d}, CamCo~\cite{Xu2024Camco}, CameraCtrl~\cite{He2024CameraCtrl}, and Wonderland~\cite{liang2024wonderland} condition pre-trained video models on poses parameterized by Pl\"ucker coordinates. 4DiM trains a pixel-based DM conditioned on both camera pose and time. ViewCrafter~\cite{Yu2024ViewCrafter} proposes a point-cloud conditioned DM, leveraging DUSt3R~\cite{Wang2024CVPRDuster} as powerful geometric prior.
While pre-trained diffusion backbones enable generalization and high image fidelity, pose-conditional DMs often struggle with generating multi-view consistent images. Some approaches address these inconsistencies with carefully tuned reconstruction algorithms~\cite{wu2023reconfusion,Gao2024Cat3d} or complicated iterative approaches~\cite{Yu2024ViewCrafter} to subsequently distill the generated images into a 3D representation. Instead, our \ourmodel directly generates a 3D representation, which can optionally be further refined with out-of-the-box reconstruction algorithms. 

\section{Method}
\label{sec:method}
\method
We introduce Generative Gaussian Splatting (\ourmodel) which directly synthesizes 3D-consistent scenes from one or more posed reference images. Key to our method is the integration of an explicit 3D representation with a pre-trained video diffusion model, which we explain in detail in~\secref{sec:3drep}. \figref{fig:method} shows an overview. In alignment with most recent works, we consider a pre-trained U-Net diffusion model~\cite{Yu2024ViewCrafter,He2024CameraCtrl,Wang2024MotionCtrl}. The video model was trained with $v$-prediction, and conditioned on a single input image by concatenation of the reference latent to the input sequence, as proposed in~\cite{Blattmann2023SVD}.

\subsection{Pose-Conditional Image-To-Video Architecture}
\label{sec:poseconddiff}
We condition the pre-trained video diffusion backbone on both camera poses and multiple reference images to enable pose-conditional view synthesis.
Consider a set of posed images $\{\mathbf{I}^m,\mathbf{p}^m\}$, comprising $K$ reference images and $L$ target images, where $K+L=M$. These images come with known camera extrinsics and intrinsics $\{\mathbf{p}^m\}$. All images are encoded into latent space. We apply the forward diffusion process only to the target views, yielding a set of noise-free reference latents $\{\mathbf{z}_{ref,0}^k\}_{k=1}^K$ and noisy target latents $\{\mathbf{z}_{tgt,t}^l\}_{l=1}^L$. To condition the model on multiple reference images, we concatenate the reference latents channelwise to the input sequence, setting their noise level to zero, \ie, $\{[\mathbf{z}_{ref,0}^k,\mathbf{z}_{ref,0}^k]\}_k$. For the target latents, we concatenate zeros $\{[\mathbf{z}_{tgt,t}^l,\mathbf{0}]\}_l$, see~\figref{fig:method}~(left).
For brevity, we denote this input sequence as $\mathbf{z}^m$. 
Utilizing the EDM-preconditioning framework~\cite{karras2022edm}, we approximate the denoising process with a neural network $D_\theta$, parameterized as follows:
\begin{equation}
\begin{split}
    D_\theta&(\mathbf{z}^m; \sigma_t, \{\mathbf{z}_{\text{ref}}^k\},\{\mathbf{p}^m\}) = c_{\text{skip}}(\sigma_t)\mathbf{z}^m\\
    & + c_{\text{out}}(\sigma_t)F_\theta\left(c_{\text{in}}(\sigma_t)\mathbf{z}^m; c_{\text{noise}}(\sigma_t), \{\mathbf{z}_{\text{ref}}^k\},\{\mathbf{p}^m\}\right), 
\end{split}
\end{equation}
with preconditioning weights $c_{\text{skip}}(\sigma_t)$, $c_{\text{out}}(\sigma_t)$, $c_{\text{in}}(\sigma_t)$, and $c_{\text{noise}}(\sigma_t)$. See \SUPP for more details. 
To incorporate camera poses into the backbone, we adopt the approach from CameraCtrl~\cite{He2024CameraCtrl}, integrating the video model with a camera encoder $\mathcal{P}$. The camera encoder processes the Pl\"ucker embeddings $\{\mathbf{P}^m\}$ of the poses $\{\mathbf{p}^m\}$ and outputs multi-scale camera embeddings, which are then used to condition the diffusion model.

\subsection{Integrating 3D Constraints}
\label{sec:3drep}
By conditioning the video backbone on multiple reference images and Plücker embeddings, our model can generate new images along a specified camera trajectory. However, it cannot directly generate 3D scenes, and the resulting images lack the consistency needed for high-quality 3D reconstruction through per-scene optimization, as shown in Table~\ref{tab:distillationSingle}.
To address this limitation, we introduce a stronger bias in the model to learn correct spatial relationships between frames. Specifically, we integrate a 3D representation that correlates features through depth-based reprojection. We choose \gs for its fast rendering and numerical stability during training.\\
A key challenge is determining where to best integrate the 3D representation within the diffusion model. The model's compressed latent space may not inherently preserve 3D consistency. Therefore, imposing 3D constraints directly within this latent space may not be effective. Our experiments confirm that predicting splats in latent space indeed does not yield good results, as seen in Table~\ref{tab:ablations}.
Instead, we integrate the 3D representation into the final upsampling block of the U-Net. Specifically, an epipolar transformer $\mathcal{T}_\text{epi}$ processes the inputs to the last upsampling block, followed by two prediction heads, $f_\phi$ and $f_g$, to generate the parameters of \gs, as illustrated in~\figref{fig:method}~(middle). Similarly to  PixelSplat~\cite{Charatan2024PixelSplat}, we use the epipolar transformer to correlate features along epipolar lines via attention. To maintain a manageable memory footprint, we only aggregate features from the two neighboring views for each input view. \\
Using $f_\phi$ and $f_g$, the splats are parameterized as
\begin{eqnarray}
    \mathbf{\mu}_k &= f_\phi(\mathbf{f}_\text{epi}) \\ 
    \left[\mathbf{\Sigma}_k,\alpha_k, \mathbf{c}_k, \mathbf{f}_k\right] &= f_g(\mathbf{f}_\text{epi}),
\end{eqnarray}
where $\mathbf{f}_\text{epi}$ denotes the feature map predicted by $\mathcal{T}_\text{epi}$, and $\mathbf{\mu}_k$, $\mathbf{\Sigma}_k$, $\alpha_k$, $\mathbf{c}_k$, $\mathbf{f}_k$ are mean, covariance, opacity, color and feature values of the $k$-th splat, respectively. 
The splats are then rendered into low resolution images $\hat{\mathbf{I}}^m_\text{LR}$ and feature maps $\mathbf{f}^m$ using
\begin{equation}
    \mathbf{f^m}(x) = \sum_{k=1}^K \mathbf{f}_k \alpha_k \mathbf{g}_{2D,k}(x; \mathbf{p}^m) \prod_{j=1}^{k-1} (1 - \alpha_j \mathbf{g}_{2D,j}(x; \mathbf{p}^m)),
\end{equation}
where $\mathbf{g}_{2D,k}(x; \mathbf{p}^m)$ is the 2D projection of the $k$-th splat to camera pose $\mathbf{p}^m$.
As shown in~\figref{fig:method}~(right), the feature maps are further refined by a block $f_v$ with skip connections to the input. $f_v$ outputs $\hat{\mathbf{v}}^m$, \ie, the weighted sum of predicted noise and sample used in the $v$-prediction objective.

\boldparagraph{Objective.} We follow the training procedure and hyperparameter choices of our internal diffusion backbone, and train the model with $v$-prediction~\cite{salimans2022progressive,karras2022edm}. To better regularize the 3D representation, we add reconstruction losses $\mathcal{L}_{LR}$ and $\mathcal{L}_{nv,LR}$ on the rendered low-resolution images of the input views $\{\hat{\mathbf{I}}^m_{LR}\}$ and $J$ additional novel viewpoints $\{\hat{\mathbf{I}}^j_{nv,LR}\}_{j=1}^J$. 
\begin{eqnarray}
    \mathcal{L}_{LR} &= \frac{1}{M} \sum_{m=1}^M || \hat{\mathbf{I}}^m_\text{LR} - \mathbf{I}^m_\text{LR}||_2^2 \\
    \mathcal{L}_{nv,LR} &=  \frac{1}{J} \sum_{j=1}^J || \hat{\mathbf{I}}^j_{nv,LR} - \mathbf{I}^j_{nv,LR}||_2^2.
\end{eqnarray}
Note that these novel views are directly rendered from the 3D representation and do not need to go through the network. Hence, this additional guidance comes at marginal computational cost while improving the results considerably, as shown in \tabref{tab:ablations}.\\
By incorporating an explicit 3D representation, our model is inherently equipped to leverage depth supervision. Consequently, when depth is available, we train with an additional depth loss, $\mathcal{L}_d$. This loss function minimizes the Euclidean distance between the predicted mean, $\mu_k$, of each per-pixel splat and its corresponding ground truth 3D coordinate. The ground truth is obtained by unprojecting the pixel coordinate $k$ with its corresponding depth $d_k$ and camera pose $\mathbf{p}^{m(k)}$
\begin{equation}
    \mathcal{L}_d = \frac{1}{K} \sum_{k=1}^k||\mu_k-\text{unproject}(d_k, \mathbf{p}^{m(k)})||_2.
\end{equation}
In PixelSplat, $f_\phi$ predicts a discrete probability density over a set of depth buckets and makes sampling differentiable by setting the opacity according to the probability of the sampled depth. However, in our setting, we find that this does not work well in conjunction with additional depth supervision. Instead, we approximate depth with a Gaussian distribution, \ie, by predicting a mean and covariance and using the reparameterization trick to differentiably sample from it. We validate this design choice in \tabref{tab:ablations}, and qualitatively observe that this significantly improves the predicted depth.

\subsection{Decoding Latent Gaussian Splats}
So far, our model can synthesize a sequence of images from a set of reference images. However, due to the refinement block $f_v$ and the decoder $\mathcal{D}$, the generated results can still become inconsistent. We address this issue by adding a 3D decoder $\mathcal{D}_{3D}$ that maps the intermediate feature maps of the epipolar transformer directly to \gs in image space.
The decoder first increases the resolution of the input feature maps with a 2D upsampler. Next, two blocks similar to $f_\phi$ and $f_g$ predict per-pixel splats $\{\hat{\mathbf{G}}^m\}$ in image space.
For memory efficiency, we freeze the diffusion model and train the 3D decoder at a fixed timestep $t=0$, \ie, we do not add noise to any of the inputs, effectively resulting in a novel view synthesis setting.
Our model can thus predict per-pixel splats to parameterize the full scene in image space. However, such a per-pixel representation highly overparameterizes the scene and does not allow for an adaptive resolution. Indeed, we observe that the generated splats are very small. To obtain a more compact representation, we optionally apply a per-scene optimization. As our goal is not to improve the method for 3D reconstruction itself, we rely on Splatfacto’s standard reconstruction algorithm for this task. We initialize the scene with the predicted splats from our model and perform 5K iterations of standard 3DGS optimization by running Splatfacto from~\cite{nerfstudio}, using the decoded images from the 2D decoder. In contrast to distillation approaches, our model only requires a few steps of per-scene optimization due to the good initialization with the predicted splats and works with out-of-the-box reconstruction algorithms without careful tuning.

\subsection{Splat Conditional Model}
\label{subsec:conditionalmodel}
While our model can be conditioned on any number of reference images, two views are generally sufficient for achieving high-quality view interpolations.
For generating scenes from more than two reference views, we can simply chain the results together. To ensure consistent content across generated batches, we condition the model on renderings of the current 3D scene, similar to the approach in~\cite{Muller2024MultiDiff}, but without relying on pre-trained monocular priors. Specifically, we utilize \gs in feature space and replace their feature values with the corresponding predicted image latents to align with our image conditioning strategy. The resulting 3D representation is rendered from the target views, and the outputs are concatenated channel-wise with the noisy image latents.
We train our conditional model with a mixture of datasets, some of which include ground truth depth. When available, we use the depth information from the reference images along with their latents to approximate the conditioning signal. We construct per-pixel splats and adjust the scale so that the re-rendered splats occupy one pixel per view. Since our model also predicts per-pixel splats of relatively small size during inference, this method of approximating the 3D representation proves effective for conditional training.

\section{Experiments}

\boldparagraph{Datasets.}
For training, we utilize three datasets of static indoor environments: \re~\cite{Zhou2018Re10k}, \sn~\cite{Dai2017Scannet}, and an internal large-scale dataset of synthetic environments. These datasets provide video sequences with registered camera parameters, and both \sn and our internal dataset include metric depth maps.
\re comprises sequences of approximately 30-100 frames from 10,000 real estate recordings, featuring smooth camera trajectories with minimal roll or pitch. Following~\cite{Watson20244DiM}, we also use a variant of \re with rescaled camera poses to be approximately metric.
\sn includes 1,513 handheld captures of indoor environments, with camera trajectories that follow a scan pattern, often exhibiting rapid changes and varied orientations. 
Our internal dataset contains around 95,000 synthetic indoor environments with smooth camera trajectories achieved through spline interpolation.
In line with previous work, we benchmark our approach on \re. Additionally, we evaluate on \snpp~\cite{Yeshwanth2023Scannetpp}, which consists of 460 indoor scenes. Despite the similar name, \snpp features different cameras and scenes from \sn, allowing us to assess the generalization capability of our method in real-world scenarios.

\boldparagraph{Experimental Settings.}
In \secref{sec:singleimg}, we consider scene synthesis from a single image. \secref{sec:twoimg} analyzes scene reconstruction and synthesis from two reference images. 
For the single-view setting, we subsample 8 frames for \re with a stride of 10, similar to ~\cite{Watson20244DiM,Yu2023PNVS}, and use a stride of 4 for \sn, since camera trajectories in \sn feature stronger motion. The first image of the sequence is assigned as reference image. During training, we randomly flip the order of the camera trajectories and
additionally sample 6 novel views between the target views, but otherwise apply the same sampling strategy for training and evaluation. We report results on 128 randomly selected scenes from the \re testset and 50 scenes from \snpp. Since the camera trajectories in \snpp can feature extremely large changes in camera motion, we select target views that have a decent overlap with the reference images, see \SUPP for details. \\
For the two-view conditional model, we comply with the training strategy of PixelSplat~\cite{Charatan2024PixelSplat} and sample 8 views randomly within a maximum gap of 80 frames for \re, and 32 frames for \sn. During training, we select the first and another randomly selected view as reference images and additionally sample 6 novel views between the target views. Given two reference images, we evaluate performance separately on view interpolation and view extrapolation. For view interpolation, the reference images are the first and last frame of the sequence, which is the common setting in regression-based NVS methods. For view extrapolation, the reference frames are the first two frames of the sequence and the model needs to extrapolate the remaining 6 views, making this a generative task.

\boldparagraph{Evaluation Metrics.}
We report metrics for image quality and 3D consistency. Peak Signal-to-Noise Ratio and LPIPS~\cite{zhang2018unreasonable} quantify reconstruction quality. To measure the overall image fidelity of the generated images, we compute FID~\cite{Heusel2017NIPS} and FVD~\cite{unterthiner2018towards}, for more details see \SUPP.
Furthermore, we evaluate the 3D consistency of our approach using the Thresholded Symmetric Epipolar Distance (TSED), which measures the alignment between pairs of views by computing the symmetric epipolar distance between corresponding points in the two views \cite{Yu2023PNVS}. We report TSED scores with a threshold of 2.0. As some of the baselines can only produce results at resolution $256\times256$ pixels, we report quantitative results at this resolution unless denoted otherwise. For \ourmodel, we report numbers on generated images using the 2D decoder.

\boldparagraph{Baselines.}
We compare \ourmodel to the strongest existing approaches that have code available. Note that all of the baselines make use of pre-trained backbones. While this enables all methods to generalize well to \snpp, it can also make a direct comparison difficult, especially when diffusion priors vary in quality. For a fair comparison of a model with and without an intermediate 3D representation, we train our own purely pose-conditional model (\ourstwod) as described in \secref{sec:poseconddiff}. Note that the video backbones of CameraCtrl\cite{He2024CameraCtrl} and ViewCrafter~\cite{Yu2024ViewCrafter} require 14, and 25 frames, respectively, while we consider an 8-frame setting. We evaluate these baselines by padding the output camera trajectory and subsequently subsampling the generated results.

\boldparagraph{Implementation Details.}
We use a proprietary pre-trained image-2-video diffusion model with $\sim2$B parameters and a U-Net architecture. The camera encoder and 3D decoder have $\sim$200M and 1.5M parameters, respectively.
All of our models were trained on 8 Nvidia A100 80GB GPUs with a batch size of 1 per GPU, using the AdamW optimizer~\cite{Loshchilov2017adamw} with a learning rate of $3\times10^{-5}$.
Starting from a pre-trained I2V diffusion model, we first finetune the temporal attention layers of the U-Net together with the camera encoder for 100K iterations. Afterwards, we freeze the parameters of the camera encoder for memory efficiency, integrate the 3D representation and, and finetune the full model.
The ablation studies are reported after training the models for 75K iterations. Our final models were trained for ~800K iterations, taking approximately 2 weeks. Inference is performed with a discrete Euler scheduler using 30 steps. 
The monocular depth predictor samples 3 \gs per pixel, unless we train with depth supervision, where we only sample a single splat. 

\subsection{Scene Synthesis From a Single Image}
\label{sec:singleimg}
\baselinecompSingle
\baselinecompfigSingle
\tabref{tab:baselinecompSingle} shows the comparison of \ourmodel against prior art for single-image to scene synthesis. On \re, our approach significantly improves image quality and 3D consistency over existing approaches. On \snpp, ViewCrafter generates images with slightly higher fidelity but less consistency as indicated by a lower TSED. Note that ViewCrafter should be compared to \ourmodel+depth, since it relies on DUSt3R~\cite{Wang2024CVPRDuster}, which was trained with ground truth depth. Interestingly, on \snpp, we observe large improvements in TSED for including a 3D representation in our model. Upon closer inspection, we find that \snpp contains trajectories that go back and forth in a scene. In these cases, \ourstwod tends to generate inconsistent content, while with 3D representation, the model can generate consistent frames. We show qualitative examples in \figref{fig:baselinecompfigSingle}. Overall, we observe that the generated images from CameraCtrl tend to be blurry in the generated areas. ViewCrafter generates visually plausible views but can alter the colors and sometimes deforms objects, see \eg the screen in the \snpp example in \figref{fig:baselinecompfigSingle}.

\subsection{Scene Synthesis From Two Images}
\label{sec:twoimg}
\baselinecompfig
We report results for two tasks: view interpolation and view extrapolation. For view interpolation, we also include the purely reconstruction based PixelSplat~\cite{Charatan2024PixelSplat} for reference in ~\tabref{tab:baselinecomp}. Both, PixelSplat and LatentSplat perform well on \re and \snpp for view interpolation. Our approach achieves similar results on \re but does not reach the same reconstruction quality on \snpp. However, PixelSplat does not support view extrapolation, which is our primary objective. The generative decoder of LatentSplat also struggles to generate realistic views far away from the reference images, reflecting in overall worse extrapolation metrics and visible artifacts, as shown in the qualitative examples in~\figref{fig:baselinecompfig}. Similar to the single image results, ViewCrafter performs particularly well on \snpp but lacks 3D consistency as indicated by a lower TSED compared to \ourmodel+depth in all settings. Qualitatively, we observe that on \snpp, ViewCrafter can struggle to match the viewpoint correctly, as visible in~\figref{fig:baselinecompfig}.\\
\baselinecomp
\distillationComp
So far, we analyzed the quality and consistency of the generated output frames of the methods. But ultimately, we are interested in generating high-quality 3D scenes. We hence conduct an additional experiment in which we reconstruct a 3D scene from the generated outputs of the strongest generative baseline, ViewCrafter, and our approach. Since we only evaluate the outputs from a single forward pass, we do not use the iterative view synthesis strategy in conjunction with a content-adaptive camera trajectory planning algorithm that ViewCrafter proposes for long-range view extrapolation. Instead, we run an off-the-shelf 3DGS optimization, \ie train a Splatfacto model from~\cite{nerfstudio} for 15,000 iterations on the multi-view images generated by ViewCrafter. 
For our method, we directly consider the generated 3D splats predicted by \ourmodel. Additionally, we train a refined variant, for which we initialize Splatfacto with the generated splats and run it for 5,000 iterations per scene to obtain a more compact 3D scene representation.
We run the per-scene optimization on all evaluation scenes and assess the quality of the resulting 3D representations by calculating FID and FVD on rendered images from viewpoints interpolated between the training views. 
The qualitative and quantitative results in ~\figref{fig:distillation} and ~\tabref{tab:distillationSingle} highlight the importance of including a 3D representation in the generative model. While the individual frames of ViewCrafter are of high quality, slight inconsistencies in the generated sequence result in clearly visible artifacts in the 3D reconstruction.
\distillationSingle

\subsection{Autoregressive Scene Synthesis}
\inpainting
To extend our model from the two-view setting to an arbitrary number of input views, we train a conditional variant to autoregressively generate a full scene, as described in~\secref{subsec:conditionalmodel}. At inference, we first generate multi-view images between the first two reference images, as well as an initial 3D representation of the scene. Next, we re-render the generated scene as condition for the next step, biasing the model towards 3D-consistent inpainting. We proceed by using the second and third reference images together with the rendered condition as input to our model, updating the 3D representation after each step. We continue with this strategy for all available reference images. 
In~\figref{fig:inpainting}, we show a reconstructed 3D scene that was generated with \ourmodel from only 5 reference images. For this example, we generate images between the reference images and further consider camera viewpoints with additional positive and negative pitch to obtain a more complete reconstruction.

\subsection{Ablation Studies}
We ablate our model choices from~\secref{sec:method} in the two-view conditional setting on \re in \tabref{tab:ablations}. Without the additional loss on novel viewpoints (No $\mathcal{L}_{nv}$), image quality and consistency drop significantly, indicated by lower PSNR and TSED values. 
Predicting the 3D representation in an intermediate feature space together with $v$-prediction clearly outperforms a 3D representation in latent space and training with $x_0$ prediction (No $f_v$, $x_0$-prediction). Compared to predicting a discrete probability distribution $\phi$ for the depths, our proposed approximation with a Gaussian distribution works better when depth supervision is available and improves results on all metrics. Lastly, we investigate two architectures for the 2D upsampler in the 3D decoder $\mathcal{D}_{3D}$: we compare a convolutional and a transformer-based upsampler, finding that for this task, the convolutional architecture performs better. Compared to the 2D decoder (GGS), the 3D decoder improves 3D consistency but moderately degrades visual fidelity.
\ablations
\section{Conclusion}
In this paper, we introduce \ourmodel, a novel approach for 3D scene synthesis from few input images. We integrate 3D representations with existing video diffusion priors to improve the consistency of the generated images and enable directly generating 3D scenes. Our experiments on RealEstate10k and ScanNet++ show that \ourmodel reduces the gap between generative models and regression-based models for view interpolation, while simultaneously achieving clear improvements over the relevant generative baselines in terms of 3D consistency and image fidelity. Nonetheless, the most recent regression-based NVS methods achieve even higher image fidelity for view interpolation. We believe that closing this gap is an important step for future work towards a holistic approach for novel view synthesis.
\noindent
{
    \small
    \bibliographystyle{ieeenat_fullname}
    \bibliography{bibliography,further_references}
}

\clearpage
\appendix
\section*{\Large\textbf{Appendix}}
We provide additional qualitative and quantitative results and discuss training and evaluation details.

\section{Background}
\subsection{Diffusion Models}
\label{app:DM}
Diffusion Models (DMs) transform data to noise by learning to sequentially denoise their inputs $\mathbf{z_T}\sim\mathcal{N}(\mathbf{0}, \sigma_T)$~\cite{sohldickstein2015deep,song2020denoising,ho2020denoising}. This approximates the
reverse ODE to a stochastic forward process which transforms the data distribution $p_\text{data}(\mathbf{z}_0)$ to an approximately Gaussian distribution $p(\mathbf{z};\sigma_T)$ by adding i.i.d. Gaussian noise with sufficiently large $\sigma_T$ and can be written as
\begin{equation}
    d\mathbf{z} = -\dot{\sigma(t)}\sigma(t)\nabla_{z}\log p(\mathbf{x};\sigma(t))dt .
\end{equation}
DM training approximates the score function $\nabla_{z}\log p(\mathbf{x};\sigma(t))$ with a neural network $\mathbf{s}_\theta(\mathbf{z};\sigma)$ with parameters $\theta$. In practice, the network can be parameterized as $\mathbf{s}_\theta(\mathbf{z};\sigma)=(D_\theta(\mathbf{z};\sigma)-\mathbf{z})/\sigma^2$ and trained via denoising score matching
\begin{equation}
    \mathbb{E}_{(z_0, y) \sim p_{\text{data}}(z_0, y), (\sigma, n) \sim p(\sigma, n)} \left[ \lambda_\sigma \left\| D_\theta(\mathbf{z}_0 + \mathbf{n}; \sigma, \mathbf{y}) - \mathbf{z}_0 \right\|_2^2 \right]
\end{equation}
where $p(\sigma, \mathbf{n}) = p(\sigma) \mathcal{N} (\mathbf{n}; \mathbf{0}, \sigma^2)$, $p(\sigma)$ is a distribution over noise levels $\sigma$, $\lambda_\sigma : \mathbb{R}^+ \rightarrow \mathbb{R}^+$ is a weighting function, and $y$ is an arbitrary conditioning signal. 
We follow the EDM-preconditioning framework~\cite{karras2022edm} and use
\begin{align*}
D_\theta(\mathbf{z}; \sigma) &= c_{\text{skip}}(\sigma)\mathbf{z} + c_{\text{out}}(\sigma)F_\theta(c_{\text{in}}(\sigma)\mathbf{z}; c_{\text{noise}}(\sigma)),
\end{align*}
where $F_\theta$ is the trained neural network and $c_\text{skip}$, $c_\text{out}$, $c_\text{in}$, and $c_\text{noise}$ are scalar weights.

\subsection{Gaussian Splatting}
In their seminal work, Kerbl \etal~\cite{kerbl3Dgaussians} propose to represent a 3D scene as a set of scaled 3D Gaussian primitives $\{\mathbf{G}_k\}_{k=1}^K$ and render an image using volume splatting. 
Each 3D Gaussian $\mathbf{G}_k$ is parameterized by an opacity $\alpha_k \in [0, 1]$, color $\mathbf{c}_k$, a center (mean) $\mathbf{p}_k \in \mathbb{R}^{3 \times 1}$ and a covariance matrix $\mathbf{\Sigma}_k \in \mathbb{R}^{3 \times 3}$ defined in world space:
\begin{equation}
    \mathbf{G}_k(\mathbf{x}) = e^{-\frac{1}{2}(\mathbf{x}-\mathbf{p}_k)^T\Sigma_k^{-1}(\mathbf{x}-\mathbf{p}_k)}
\end{equation}
In practice, the covariance matrix is calculated from a predicted scaling vector $s \in \mathbb{R}^3$ and a rotation matrix $\mathbf{O} \in \mathbb{R}^{3 \times 3}$ to constrain it to the space of valid covariance matrices, \ie,
\begin{equation}
    \mathbf{\Sigma}_k = \mathbf{O}_k \mathbf{s}_k \mathbf{s}_k^T \mathbf{O}_k^T.
\end{equation}
The color $c_{k}$ is parameterized with spherical harmonics to model view-dependent effects. 
To render this 3D representation from a given viewpoint with camera rotation $\mathbf{R} \in \mathbb{R}^{3 \times 3}$ and translation $\mathbf{t} \in \mathbb{R}^3$, the Gaussians $\{G_k\}$ are first transformed into camera coordinates
\begin{equation}
\mathbf{p}_k' = \mathbf{R} \mathbf{p}_k + \mathbf{t}, \quad \mathbf{\Sigma}_k' = \mathbf{R} \mathbf{\Sigma}_k \mathbf{R}^T
\end{equation}
and susequently projected to ray space using a local affine transformation
\begin{equation}
    \mathbf{\Sigma}_k'' = \mathbf{J}_k \mathbf{\Sigma}_k' \mathbf{J}_k^T,
\end{equation}
where the Jacobian matrix $\mathbf{J}_k$ is an affine approximation to the projective transformation defined by the center of the 3D Gaussian $\mathbf{p}_k'$.
The Gaussians are projected onto a plane by skipping the third row and column of $\mathbf{\Sigma}_k''$, yielding the 2D covariance matrix $\Sigma_{2D,k}$ of the projected 2D Gaussian $\mathbf{G}_{2D,k}$.
The rendered color is obtained via alpha blending according to the primitive's depth order $1, \ldots, K$:
\begin{equation}
\mathbf{c}(x) = \sum_{k=1}^K \mathbf{c}_k \alpha_k \mathbf{G}_{2D,k}(x) \prod_{j=1}^{k-1} (1 - \alpha_j \mathbf{G}_{2D,j}(x)).
\end{equation}

\section{Implementation Details}
\boldparagraph{Model architecture.} 
Our epipolar transformer consists of 2 attention blocks with 4 heads each. Similar to~\cite{Charatan2024PixelSplat}, it samples 32 values per feature map on each pixel’s epipolar line. As the transformer already operates on a low-resolution latent space, we do not apply any spatial downsampling. The depth predictor consists of 2 linear layers with ReLU and sigmoid activation that predict mean and variance of the per-pixel disparity. The disparity is further mapped to an opacity with 4 additional linear layers with ReLU activation, followed by a sigmoid activation. In parallel, the remaining Gaussian parameters, \ie, scale, rotation, color and feature values, as well as a per-pixel offset are predicted with a linear layer from the feature maps predicted by the epipolar transformer. For efficiency, we use a 32 channels for the feature values per Gaussian and omit view-dependent effects, \ie, predict rgb values instead of spherical harmonics. \\
The architecture of the 3D decoder consists of a 2D upsampler and architecturally similar layers to the aforementioned depth predictor and mapping to Gaussian parameters. The 2D upsampler consists of multiple blocks 2D convolutions with replication padding and nearest neighbor upsampling, resulting in a total of $1.5$M parameters for the 3D decoder. 

\section{Baselines}
\boldparagraph{PNVS}
We use the official implementation of the authors \url{https://github.com/YorkUCVIL/Photoconsistent-NVS.git} and the provided checkpoint on \re. \\
\boldparagraph{MultiDiff}
We run code and a checkpoint for \re, both provided by the authors, using our evaluation splits. \\
\boldparagraph{CameraCtrl}
We use the official implementation of the authors \url{https://github.com/hehao13/CameraCtrl.git} and the provided checkpoint on \re. Since the original implementation is trained to generate 14 frames, we pad the camera trajectory by duplicating frames and subsequently subsampling the generated images.\\
\boldparagraph{ViewCrafter}
We use the official evaluation scripts provided by the authors \url{https://github.com/Drexubery/ViewCrafter.git}. Since the original implementation is trained to generate 25 frames, we pad the camera trajectory by duplicating frames and subsequently subsampling the generated images.
\\
\boldparagraph{PixelSplat}
We run the official \re-checkpoint and inference implementation \url{https://github.com/dcharatan/pixelsplat.git} using our evaluation splits. We remark that the results on our evaluation split are lower than the originally reported numbers in~\cite{Charatan2024PixelSplat} on the full testset. Since evaluating generative methods on such a large quantity of scenes is computationally expensive and slow, we decided to only report numbers on 128 randomly sampled scenes from the testset, following~\cite{Watson20244DiM}. We verified that we obtain the originally reported performance when using their original evaluation split to ensure we run the method correctly and note that another work also measured lower performance for PixelSplat on a slightly different evaluation split~\cite{Wewer2024latentsplat}.\\
\boldparagraph{LatentSplat}
We evaluate the official checkpoint on \re using the official inference implementation \url{https://github.com/Chrixtar/latentsplat.git} together with our evaluation splits.\\
\boldparagraph{4DiM}
While we designed our evaluation setting for \re approximately similar to 4DiM~\cite{Watson20244DiM}, a quantitative comparison is difficult because no official code or evaluation splits are available. We observe that reconstruction quality on \re can vary significantly between scenes, as indicated by a large standard deviation for both reconstruction metrics: $19.2\pm4.2$ for PSNR and $0.277\pm0.113$ for LPIPS in our single-view setting with 128 randomly sampled scenes. For reference, 4DiM reports a PSNR of $18.09$ and LPIPS of $0.263$ for their best model. We also measure a slightly lower TSED when using ground truth data: $0.993$ whereas 4DiM obtains $1.000$ on their evaluation split. Note that a TSED below 1.0 can indeed happen on ground truth data, as poses in the data are noisy and do not achieve a perfect score~\cite{Watson20244DiM}.
Considering the results for our approach $0.992$($0.993$) and $0.997$($1.000$) for 4DiM, both methods saturate the metric \wrt the corresponding evaluation split. Lastly, we remark that FID and FVD depend strongly on the number of real samples that were used for comparison, as well as any preprocessing of the data. 4DiM does not provide these evaluation details and their numbers for FID and FVD are not directly comparable with our results. 

\section{Experimental Setting}
\label{sec:supp_baselines}
\subsection{ScanNet++}
The camera trajectories for \snpp follow a scan-pattern and viewpoints often change rapidly with large camera motion. 
When evaluating our method, we hence ensure that the sampled target views have an average overlap of at least 50\% with the reference views. Specifically, we sample one or two reference views randomly and then compute the overlap for each view in the scene with the reference views using the provided ground truth depth. The overlap score of each view is computed as the average score over all reference frames. We select the views with the largest overlap as target views and discard cases in which any of the selected target views has less than 50\% average overlap.

\subsection{Metrics}
For FID, we take all generated views and compare their distribution to the same number of views for 20K scenes of the training set for \re and 171 scenes for \snpp.
For FVD, we use all generated views and the same number of views from 2048 \re scenes and 171 scenes from \snpp. Since the feature extractor for FVD requires a minimum number of 9 frames, we use reflection padding to pad real and generated sequences.\\
We use a guidance scale of 2.0 for all our results.

\section{Additional Results}

\subsection{Teaser Images}
The images in Fig.~1 of the main paper show the generated splats with a subsequent per-scene optimization, running the default Splatfacto method of \cite{nerfstudio} with 5,000 iterations. To encourage a fewer splats, we initialize the scale based on the average distance to the three nearest neighbors instead of directly using the predicted scale. We also visualize the generated splats in feature space, \ie, the 3D representation generated by our model. Since the features are high-dimensional, we show the first three principal components of feature space. We use the same visualization for the images shown in this \SUPP.

\subsection{Additional Quantitative Results}
\distillation
\ablationsApp
We provide additional quantitative results for 3D scene synthesis using two reference images in~\tabref{tab:distillation}. We further conduct an additional ablations study, for which we only train a 3D decoder on top of a frozen diffusion model, \ie, Ours-No3D. As shown in ~\tabref{tab:ablationsApp}, this approach consistently performs worse than GGS and GGS with 3D decoder, corroborating our design choice to include the 3D representation directly in the diffusion model to synthesize consistent results.

\subsection{Additional Qualitative Results}
\baselinecompfigSingleSupp
\baselinecompfigSupp
\singleViewScene
\distillationCompSupp
\inpaintingSupp
We show additional baseline comparisons for synthesis from a single view 
and two views in \figref{fig:baselinecompfigSingleSupp} and \figref{fig:baselinecompfigSupp}, respectively. \figref{fig:singleViewScene} provides more generated 3D scenes from a single image, using our \ourmodel model and \figref{fig:distillationCompSupp} depicts additional comparisons to ViewCrafter~\cite{Yu2024ViewCrafter} for 3D scene synthesis. 
Lastly, we include more autoregressive scene synthesis results in \figref{fig:inpaintingSupp}

\end{document}